\documentclass[11pt,a4paper]{article}
\usepackage{times,latexsym}
\usepackage{url}
\usepackage[T1]{fontenc}

\usepackage[acceptedWithA]{tacl2021v1}
\usepackage{xspace,mfirstuc,tabulary}

\newif\iftaclinstructions
\taclinstructionsfalse % AUTHORS: do NOT set this to true
\iftaclinstructions
\renewcommand{\confidential}{}
\renewcommand{\anonsubtext}{(No author info supplied here, for consistency with
TACL-submission anonymization requirements)}
\newcommand{\instr}
\fi

\iftaclpubformat % this "if" is set by the choice of options

\else

\fi

\usepackage{microtype}
\usepackage{booktabs}
\usepackage{makecell}
\usepackage{graphicx}
\usepackage{adjustbox}
\usepackage{xspace}
\usepackage{pifont}
\usepackage{xcolor}
\usepackage{color, colortbl}
\usepackage{multicol}
\usepackage{multirow}
\usepackage{amsmath}
\usepackage{wasysym}
\usepackage[capitalise]{cleveref}
\definecolor{burgundy}{rgb}{0.5, 0.0, 0.13}
\definecolor{cerise}{rgb}{0.87, 0.19, 0.39}
\definecolor{cadmiumred}{rgb}{0.89, 0.0, 0.13}
\definecolor{brightmaroon}{rgb}{0.76, 0.13, 0.28}
\definecolor{bostonuniversityred}{rgb}{0.8, 0.0, 0.0}
\definecolor{jazzberryjam}{rgb}{0.65, 0.04, 0.37}

\newcommand{\checkcolor}[1]{\textcolor{blue}{#1}}
\newcommand{\crosscolor}[1]{\textcolor{red}{#1}}
\definecolor{tablegrey}{rgb}{0.93, 0.93, 0.93}

\newcommand{\twod}{2D\xspace}
\newcommand{\threed}{3D\xspace}
\newcommand{\gridluds}{gridlu\xspace}
\newcommand{\minigridds}{minigrid\xspace}
\newcommand{\minedojods}{minedojo\xspace}
\newcommand{\minerlds}{minerl\xspace}
\newcommand{\alfredds}{alfred\xspace}
\newcommand{\embodiedqads}{embodiedqa\xspace}
\newcommand{\alfworldds}{alfworld\xspace}
\newcommand{\scienceworldds}{scienceworld\xspace}
\newcommand{\jerichods}{jericho$_{\textsc{agentbench}}$\xspace}
\newcommand{\blocksworldds}{blocksworld$_{\textsc{apbench}}$\xspace}
\newcommand{\blocksworldshortds}{blocksworld\xspace}
\newcommand{\goldminerds}{goldminer$_{\textsc{apbench}}$\xspace}
\newcommand{\hotpotds}{hotpotqa$_{\textsc{react}}$\xspace}
\newcommand{\gsmeightkds}{gsm8k$_{\textsc{toolqa}}$\xspace}
\newcommand{\gqads}{gqa\xspace}
\newcommand{\imgeditds}{imgedit\xspace}
\newcommand{\gaiads}{gaia\xspace}
\newcommand{\mandmds}{m\&m's\xspace}
\newcommand{\toolbenchds}{toolbench\xspace}
\newcommand{\restbenchds}{restbench\xspace}
\newcommand{\toolalpacads}{toolalpaca\xspace}
\newcommand{\mindtowebds}{mind2web\xspace}
\newcommand{\aitwds}{aitw\xspace}
\newcommand{\osworldds}{osworld\xspace}
\newcommand{\appworldds}{appworld\xspace}
\newcommand{\databaseds}{sql\_database$_{\textsc{agentbench}}$\xspace}

\newcommand{\llfbench}{llf-bench\xspace}
\newcommand{\mint}{mint\xspace}

\newcommand{\agentboard}{agentboard\xspace}

\newcommand{\coloredhref}[2]{\hyperref[#1]{\textcolor{black}{#2}}}
\newcommand{\buildanchor}[2]{\phantomsection\label{#1}#2}

\newcommand{\initgridlu}{\buildanchor{gridlu}{\gridluds}}
\newcommand{\gridlu}{\coloredhref{gridlu}{\gridluds}}
\newcommand{\initminigrid}{\buildanchor{minigrid}{\minigridds}}
\newcommand{\minigrid}{\coloredhref{minigrid}{\minigridds}}
\newcommand{\initminerl}{\buildanchor{minerl}{\minerlds}}
\newcommand{\minerl}{\coloredhref{minerl}{\minerlds}}
\newcommand{\initminedojo}{\buildanchor{minedojo}{\minedojods}}
\newcommand{\minedojo}{\coloredhref{minedojo}{\minedojods}}
\newcommand{\initalfred}{\buildanchor{alfred}{\alfredds}}
\newcommand{\alfred}{\coloredhref{alfred}{\alfredds}}
\newcommand{\initembodiedqa}{\buildanchor{embodiedqa}{\embodiedqads}}
\newcommand{\embodiedqa}{\coloredhref{embodiedqa}{\embodiedqads}}
\newcommand{\initalfworld}{\buildanchor{alfworld}{\alfworldds}}
\newcommand{\alfworld}{\coloredhref{alfworld}{\alfworldds}}
\newcommand{\initscienceworld}{\buildanchor{scienceworld}{\scienceworldds}}
\newcommand{\scienceworld}{\coloredhref{scienceworld}{\scienceworldds}}
\newcommand{\initjericho}{\buildanchor{jericho}{\jerichods}}
\newcommand{\jericho}{\coloredhref{jericho}{\jerichods}}
\newcommand{\initblocksworld}{\buildanchor{blocksworld}{\blocksworldds}}
\newcommand{\blocksworld}{\coloredhref{blocksworld}{\blocksworldds}}
\newcommand{\blocksworldshort}{\coloredhref{blocksworld}{\blocksworldshortds}}
\newcommand{\initgoldminer}{\buildanchor{goldminer}{\goldminerds}}
\newcommand{\goldminer}{\coloredhref{goldminer}{\goldminerds}}
\newcommand{\inithotpot}{\buildanchor{hotpot}{\hotpotds}}
\newcommand{\hotpot}{\coloredhref{hotpot}{\hotpotds}}
\newcommand{\initgsmeightk}{\buildanchor{gsmeightk}{\gsmeightkds}}
\newcommand{\gsmeightk}{\coloredhref{gsmeightk}{\gsmeightkds}}
\newcommand{\initgqa}{\buildanchor{gqa}{\gqads}}
\newcommand{\gqa}{\coloredhref{gqa}{\gqads}}
\newcommand{\initimgedit}{\buildanchor{imgedit}{\imgeditds}}
\newcommand{\imgedit}{\coloredhref{imgedit}{\imgeditds}}
\newcommand{\initgaia}{\buildanchor{gaia}{\gaiads}}
\newcommand{\gaia}{\coloredhref{gaia}{\gaiads}}
\newcommand{\initmandm}{\buildanchor{mandm}{\mandmds}}
\newcommand{\mandm}{\coloredhref{mandm}{\mandmds}}
\newcommand{\inittoolbench}{\buildanchor{toolbench}{\toolbenchds}}
\newcommand{\toolbench}{\coloredhref{toolbench}{\toolbenchds}}
\newcommand{\initrestbench}{\buildanchor{restbench}{\restbenchds}}
\newcommand{\restbench}{\coloredhref{restbench}{\restbenchds}}
\newcommand{\inittoolalpaca}{\buildanchor{toolalpaca}{\toolalpacads}}
\newcommand{\toolalpaca}{\coloredhref{toolalpaca}{\toolalpacads}}
\newcommand{\initmindtoweb}{\buildanchor{mindtoweb}{\mindtowebds}}
\newcommand{\mindtoweb}{\coloredhref{mindtoweb}{\mindtowebds}}
\newcommand{\initaitw}{\buildanchor{aitw}{\aitwds}}
\newcommand{\aitw}{\coloredhref{aitw}{\aitwds}}
\newcommand{\initosworld}{\buildanchor{osworld}{\osworldds}}
\newcommand{\osworld}{\coloredhref{osworld}{\osworldds}}
\newcommand{\initappworld}{\buildanchor{appworld}{\appworldds}}
\newcommand{\appworld}{\coloredhref{appworld}{\appworldds}}
\newcommand{\initdatabase}{\buildanchor{database}{\databaseds}}
\newcommand{\database}{\coloredhref{database}{\databaseds}}

\newcommand{\tfont}[1]{\textsc{#1}} % for tasknames
\newcommand{\exfont}[1]{\textit{#1}} % for example
\newcommand{\exfonttable}[1]{\textit{#1}} % for examples in table
\newcommand{\afont}[1]{\texttt{#1}} % for actions
\newcommand{\tfontintro}[1]{\textbf{#1}} % for task names in intro

\newcommand{\worldstate}{world state\xspace}
\newcommand{\worldstates}{world states\xspace}

\newcommand{\kstates}{knowledge states\xspace}

\title{A Survey on Complex Tasks for Goal-Directed Interactive Agents}

\author{
  Mareike Hartmann and Alexander Koller
  \\
  Department of Language Science and Technology
  \\
  Saarland Informatics Campus
  \\
  Saarland University, Saarbrücken, Germany
  \\
  \texttt{\{mareikeh, koller\}@coli.uni-saarland.de}
}

\date{}

\begin{document}
\maketitle
\begin{abstract}
Goal-directed interactive agents, which autonomously complete tasks through interactions with their environment, can assist humans in various domains of their daily lives. Recent advances in large language models (LLMs) led to a surge of new, more and more challenging tasks to evaluate such agents. To properly contextualize performance across these tasks, it is imperative to understand the different challenges they pose to agents. To this end, this survey compiles relevant tasks and environments for evaluating goal-directed interactive agents, structuring them along dimensions relevant for understanding current obstacles. An up-to-date compilation of relevant resources can be found on our project website: \url{https://coli-saar.github.io/interactive-agents}.
%Important to understand the property of these tasks/datasets, to inform the class of approaches we can use for solving them, to contextualize results (where are we?), to understand what specific properties cause problems.
%While several surveys exist on approaches for modeling the agents, none of them gives a detailed overview over different tasks and datasets. This is what we are doing here.
\end{abstract}

%TODOS
% add wu et al counterfactuals

% !TeX root = main.tex

\section{Introduction}\label{sec:introduction}

\begin{figure*}[ht]
    \centering
    \includegraphics[width=\linewidth]{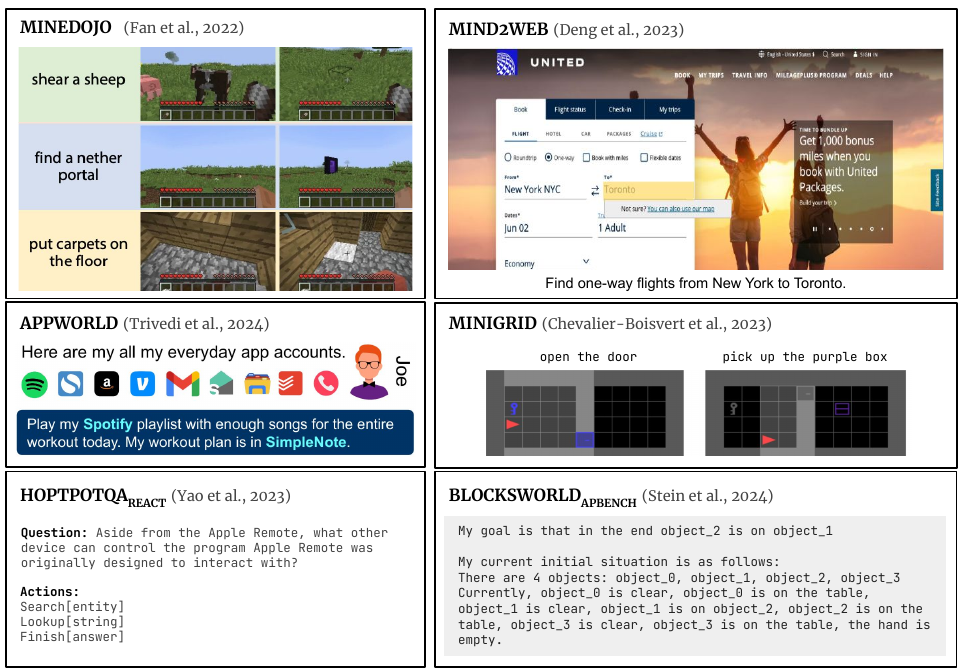}
    \caption{Examples illustrating the diversity of agent tasks. Each tasks comes with different instructions, different situations, and different possible actions.}
    \label{fig:examples}
\end{figure*}

The flurry of recent work on LLMs and tool use promises to fundamentally change the way that humans interact with computers. In the past, users had to spell out the computer's actions one by one, e.g.\ by issuing commands or clicking on GUI elements. The hope is that future users can delegate a high-level task to the computer, and it is the computer's job to decompose it into commands and actions that can be directly executed. There has been rapid progress towards agents that solve such tasks, e.g.\ by improving the ability of LLM agents to reason over contexts \cite{cot,react}, decompose problems \cite{adapt}, make decisions on what tools to use \cite{toolformer} or what actions to take \cite{li2022pre}, mostly based on in-context learning with very large LLMs \cite{brown2020language}.

At the same time, there has been intense research on the development of \emph{tasks} on which these agents can be evaluated. Such tasks range from managing our email conversations with friends (\tfont{\appworld}, \citealp{appworld}), answering complex questions (\tfont{\hotpot}, \citealp{hotpot}), doing our online shopping (\tfont{\mindtoweb}, \citealp{mind2web}), and performing complex tasks in situated environments (\tfont{\minedojo}, \citealp{minedojo}); 
see \cref{fig:examples} for some examples.
The increasing breadth, naturalness, and difficulty of these tasks has driven both a clearer understanding of the abilities and limitations of agents and pushed the development of ever more capable agent architectures.

However, it is also becoming increasingly hard to keep track of these tasks, to 
interpret the results of experiments on each task correctly, and to understand the specific challenges that each task poses for an agent. In this paper, we offer a \textbf{survey of the current landscape of tasks for goal-directed agents} that interact with their task environments. This complements existing surveys on the current landscape of agent architectures for such tasks \cite{agentbench, agentboard, gentopia, opendevin}. We hope to keep our task survey up to date in the face of the rapid current developments through a companion website\footnote{\url{https://coli-saar.github.io/interactive-agents}}, to which task developers can contribute through pull requests.

We have structured the task landscape along a number of dimensions. 
Depending on the modality of the environment (such as simulated physical environments, websites, and databases), action spaces can be as different as navigating the physical world, controlling a mouse and keyboard, selecting HTML elements on a web page, or generating SQL commands. %SHORT These action spaces  differ greatly in size and complexity.
At the same time, tasks also differ in the observability of the environment, the structure of the rewards, and the evaluation metrics. These properties greatly affect the modeling choices for a successful agent, and we hope that our survey %(with its companion website) 
will facilitate the future %SHORT evaluations and  
development of agents.

After specifying the exact scope of this survey in Section~\ref{sec:scope}, we will stake out the space of tasks we consider here through some examples in Section~\ref{sec:datasets}. We will then introduce and discuss some structuring dimensions for goal-oriented tasks in Section~\ref{sec:tasks}. In Section~\ref{sec:discussion}, we discuss our findings and offer some thoughts on future directions.

\section{Scope of this Survey}\label{sec:scope} 

Our survey comprises tasks for goal-directed interactive agents. %SHORT We focus on virtual agents rather than embodied robots (e.g. \citealp{ahn2022i}). 
By \emph{goal-directedness}, we mean that agents receive explicit goal specifications (e.g. in the form of a natural language (NL) instruction or question), from which information on goal conditions can be derived. For example, in a chess environment, we can derive goal conditions from the instruction \exfont{Checkmate the king}, whereas instructions like \exfont{Win the game} do not provide any information about goal conditions. Consequently, we do not include tasks specifying goal conditions via reward functions only (e.g. \citealp{nethack, civrealm}).

We focus on challenging tasks that require agents to map a goal to a sequence of multiple actions, rather than tasks with step-by-step instructions with a one-to-one correspondence between a part of the instruction and a necessary action (e.g. \citealp{pixelhelp, russ}). Finally, we delimit the scope to tasks solvable by a single autonomous agent without requiring a human-in-the-loop, and will discuss extensions to this paradigm in \cref{sec:discussion}, including tasks requiring interaction with a human (e.g. \citealp{lin2024decisionoriented,lateval}), and tasks requiring collaboration between multiple agents (e.g. \citealp{sotopia, gammabench}).%, and tasks requiring an agent to interact with a human (e.g. \citealp{lin2024decisionoriented,lateval}).

%Our survey focuses on tasks and applications for virtual goal-directed agents, and does not include tasks for embodied robots (e.g. \citet{ahn2022i}). %We focus on multi-step tasks, which require the agent to take a sequence of actions in order to complete the task \mareike{exclude tasks like HumanEval requiring generation of one line of code?}. Further, we only consider tasks with explicit goal specifications (e.g. in the form of a natural language instruction or question), from which information on goal conditions can be derived (e.g. in a chess environment, we can derive goal conditions from the instruction \exfont{Checkmate the king}, and would not consider instructions like \exfont{Win the game}, which does not provide any information about goal conditions), and do not include tasks specifying goal conditions via reward functions only (e.g. \cite{nethack, civrealm}). 

%We do not include tasks with step-by-step instructions, with a one-to-one correspondence between a part of the instruction and a necessary action (e.g. \citealp{pixelhelp, russ}). 
%We focus on single agent tasks, and do not include tasks requiring collaboration or interactions between multiple agents (e.g. \citealp{sotopia, gammabench}). Finally, we do not consider goal-directed tasks requiring an agent to interact with a human (e.g. \citealp{lin2024decisionoriented,lateval}), which we discuss as an interesting extension for goal-directed tasks in \cref{sec:discussion}.

\paragraph{Related surveys}
Several recent surveys on LLM-based agents exist. %\footnote{Existing surveys on specific types of environments or specific agent applications will be mentioned in the respective parts of \cref{sec:applications}.} 
\citet{mialon2023augmented}, \citet{qin2023tool}, \citet{xi2023rise}, \citet{gao2023large}, \citet{interactivenlp}, and \citet{cheng2024exploring} provide comprehensive overviews over the single agent paradigm, with a focus on modeling aspects, e.g. implementations of different components in an agent system, and common applications.  \citet{zhang2024surveymemorymechanismlarge} survey memory components for LLM-based agents. \citet{xie2024large} focus on agents based on multi-modal foundation models, \citet{guo2024large}, \citet{sun2024llmbased}, and \citet{zhang2024llm} focus on the paradigm of multi-agent interaction, \citet{xu2024survey} on the application of game-playing agents. \citet{wang2024tools} provide a detailed survey on the concept of \textit{tools} for augmenting LLMs, including a detailed overview over benchmarks for tool/API-use. \citet{peng2024survey} include an overview of a broad range of applications for LLM-based agents as one part in a pipeline for holistic evaluation of LLMs. These works mainly focus on modeling aspects and general applications, whereas our focus is on concrete tasks and environments used to develop and evaluate such agents.

%focus on methods and modeling aspects, and none of them provides a detailed description and comparison of the broad range of actual environments and tasks solved by current state-of-the-art agents.

Surveys by \citet{luketina2019survey}, \citet{madureira2020an} and \citet{cao2024survey} focus on the intersection of reinforcement learning and language-based environment representations. \citet{ijcai2022p770} survey methods for goal-conditioned reinforcement learning, solving the same type of tasks as focused on in this survey.  %The existing surveys focus on methods and modeling aspects, and none of them provides a detailed description and comparison of the broad range of actual tasks solved by current state-of-the-art agents, which is the main contribution of our survey.
\section{Example Tasks}\label{sec:datasets}

%Our literature search was based on citation graphs of works in relevant conferences and on arXiv, followed by manual filtering of candidate papers according to the criteria specified in \cref{sec:introduction}, yielding a pool of \mareike{\#}papers (at the time of writing). In the following section, we introduce a representative subset of these tasks (the tasks included in \cref{ t:detailed }), which will serve as examples to illustrate and discuss different task properties in \cref{sec:tasks}.
In the following, we introduce a set of agent tasks that will allow us to illustrate a range of different task characteristics in \cref{sec:tasks}. The full set of surveyed works can be found on our project website.
%\mareike{Say sth about literature search and filtering?}

\subsection{Navigation \& Object Manipulation in Simulations of Physical Worlds}\label{sec:objectmanipulation}
Agents can navigate and interact with physical objects in more or less realistic simulations of physical environments described in visual or textual form.\footnote{A detailed comparison between text, \twod , and \threed worlds can be found in \citet{jansen2022systematic}.}

\paragraph{Visual worlds}
\twod worlds are usually based on a \twod grid layout, %i.e. a \twod plane with cells in discretized adjacent positions, 
and observations correspond to top-down views on this grid. \tfontintro{\tfont{\initgridlu}}\footnote{We use small capitals to refer to agent tasks. Such tasks are often derived from existing datasets, e.g by adding an action space, or annotations of goal specifications. We indicate this by a subscript, e.g. \tfont{A$_\tfont{B}$} refers to a task derived from the original dataset \tfont{A} by the paper/benchmark/work \tfont{B}.} \cite{gridlu} requires an agent to put objects referred to by shapes and colors in specific spatial relations (\exfont{west of}, \exfont{north of}, ...) or spatial arrangements (\exfont{diagonal line}, \exfont{circle}, ...).
 \tfontintro{\tfont{\initminigrid}} \cite{minigrid} involves \twod grid navigation and collection of objects, e.g. picking up keys required to pass trough doors. 
 
 \textbf{\tfont{\initminerl}} \cite{basalt2021} and \textbf{\tfont{\initminedojo}} \cite{minedojo} tasks place agents in a Minecraft environment\footnote{\url{https://www.minecraft.net}}, a \threed game environment with a block-based, pixelated world representation, enabling agents with ego-centric vision to harvest materials, craft items, and build structures.
\tfontintro{\tfont{\initalfred}} \cite{alfred} places agents in a kitchen scene in the photo-realistic \threed AI2-Thor environment \cite{ai2thor}, and requires completion of typical household tasks like washing an apple in the sink. \tfontintro{\tfont{\initembodiedqa}} \cite{embodiedqa} requires agents to navigate a \threed environment in order to answer questions about rooms and objects (e.g. \exfont{Is there a bathtub in the bathroom?}).

%\threed worlds range from voxel-based state representations such as the Minecraft environments to more realistic simulations of the world with various physics simulators such as AI2-Thor as underlying engine for the \tfont{\alfred} environment. %Tasks require state changes and completion is assessed based on correspondence between final and gold states. 

%\begin{itemize}
%\item Visual worlds come with observations in \twod or \threed spaces. \twod worlds are usually based on a \twod grid layout, i.e. a \twod plane with cells in discretized adjacent positions. \threed worlds range from voxel-based state representations such as the Minecraft environments to more realistic simulations of the world with various physics simulators such as AI2-Thor as underlying engine for the \tfont{\alfred} environment. 
%\item Tasks usually require state changes and completion is assessed based on correspondence between final and gold states. 
%\end{itemize}

\paragraph{Text worlds}
 Text worlds \cite{textgames, textworld, jansen2022systematic} are environments represented via textual descriptions. %naturally fall into the intersection between NLP and sequential decision-making: they are virtual environments represented via textual descriptions. %The existence of flexible problem generators \cite{textworld} makes them interesting for targeted studies of specific challenges. % (constraints on inventory size, long-range dependencies, i.e. prerequisites for solving a task have to be completed at earlier point in time, time-based events, i.e. environment state changes not triggered by agent actions).
\tfontintro{\tfont{\initalfworld}} \cite{alfworld} transfers the \tfont{\alfred} tasks to a text world provided by the TextWorld engine \cite{textworld}.
\tfontintro{\tfont{\initscienceworld}} \cite{scienceworld} requires agents to perform scientific experiments in an environment with realistic simulations of physical, chemical, and biological processes.

\tfontintro{\tfont{\initjericho}} \cite{agentbench} is based on a subset of text games in Jericho \cite{jericho}, an environment supporting text-based fiction games such as Zork. Such games often do not include a concrete goal specification, but provide a partial description of the environment (e.g. the room the agent is located in), and some event the agent needs to react to (e.g. ringing of a telephone). Agent actions trigger new events, and winning the game requires exploring different options for what to do. \citet{agentbench} transform parts of these games into goal-directed tasks, by annotating them with concrete goals (e.g. \exfont{You need to get out of your cell out into the corridor without getting caught}).

\paragraph{PDDL planning problems}\label{sec:pddl}
These types of tasks are toy problems specified in the Planning Domain Description Language (PDDL), explicitly specifying action preconditions and effects. Compared to other tasks, action and state spaces are small and usually fully observable. The \tfontintro{\tfont{\initblocksworld}} domain requires agents to arrange different blocks into a specific configuration, usually one or more vertical stacks. The \tfontintro{\tfont{\initgoldminer}} domain places an agent in a mine, with the goal to expose gold by using bombs and laser to destroy stones. Current LLM-based agents either directly consume PDDL statements \cite{silver2022pddl}, or verbalized domain descriptions \cite{valmeekam2022large}. Entries in \cref{ t:detailed } are based on verbalizations provided  by the AutoPlanBench framework \cite{autoplanbench}.
%\begin{itemize}
%    \item These types of tasks are specified in the Planning Domain Description Language (PDDL), explicitly specifying action action preconditions and effects.
%    \item For PDDL problems, search-based planner exist
%    \item Compared to other tasks, action and state spaces are small and usually fully observable. 
%    \item LLM-based agents are applied to PDDL representations directly \mareike{CITE}, or verbalized versions of the environment and observations \cite{autoplanbench}.
%    \item This class of tasks is interesting to investigate the following problems:
%end{itemize}

\subsection{Digital Assistance}\label{sec:digitalassistants}
 Agents in digital worlds assist users with day-to-day tasks %, and usually solve real-world problems directly useful for end users. 
 comprising a vast range of applications, %like retrieving information, e.g. looking up information in databases, day-to-day tasks like booking travel or sending messages, analysing data, e.g. tables or images as in table or visual QA, and processing data, e.g. image editing. Some recent tasks focus on assistance in specific domains, e.g. the discovery of new chemicals \cite{chemcrow}, or fixing software bugs \cite{swebench}.\mareike{change cit.}. 
 like travel booking, sending emails, analysing tables or images, and editing images. Some recent tasks focus on assistance in specific domains, e.g. the discovery of new chemicals \cite{chemcrow}, or fixing software bugs \cite{swebench}.%Agents can achieve these tasks in different ways: by interacting with external pieces of software (tools or APIs, \cref{sec:apis}), by interacting with graphical user interfaces (GUIs, \cref{sec:gui}), or by interacting with code interpreters (\cref{sec:programming}).
%\begin{itemize}
%    \item Agents in digital worlds assisting users with day-to-day tasks, and usually solve real-world problems directly useful for end users. This comprises a vast range of application, like retrieving information, e.g. looking up information in databases, day-to-day tasks like booking travel or sending messages, analysing data, e.g. tables or images as in table or visual QA, and processing data, e.g. image editing. Some recent tasks focus on assistance in specific domains, e.g. the discovery of new chemicals \cite{chemcrow}, or fixing software bugs \cite{swebench}.\mareike{change cit.}
%    \item Task differ in how the agents can achieve these tasks: by interacting with external pieces of software (tools or APIs), interacting with graphical user interfaces, interacting with code interpreters
%\end{itemize}
\subsubsection{Interaction with Tools and APIs}\label{sec:apis}
Digital assistants can solve tasks by operating external pieces of software via Application Programming Interfaces (API)s, also referred to as tools\footnote{We consider APIs a special type of tools, with more complex functionality, and use both expressions interchangeably.}. Action spaces correspond to valid tool calls. APIs can be called in isolation, e.g. by indicating API name and arguments, or embedded in code (see \cref{sec:programming}). %Observations usually correspond to the APIs return values in case execution succeeds. In case of execution failure, information about error causes can be encoded into the error messages of exceptions thrown by the APIs directly (e.g. incorrect date formatting). 
%Several QA datasets have been re-purposed to agent tasks by providing a set to tools or functions the agent can use: 

\tfontintro{\tfont{\inithotpot}} \cite{react} extends the HotPotQA dataset \cite{hotpot} for multi-hop QA, which grounds answers in multiple Wikipedia documents. %, i.e. question answering requires retrieving and reasoning over multiple Wikipedia documents. 
The dataset is extended with an action space consisting of a \afont{search} action for searching for an entity's Wikipedia page, a \afont{lookup} action for strings in an article, and a \afont{finish} action to submit a final answer. \tfontintro{\tfont{\initgsmeightk}} \cite{toolqa} consists of a subset of questions from the GSM8K dataset \cite{gsm8k} for numerical reasoning over an input text %, which a state-of-the-art LLM is unable to answer, 
extended with 13 actions, e.g. for calculating the value of an equation, for executing python code, and for submitting a final answer. 

\tfontintro{\tfont{\initgqa}} \cite{visualprogramming} extends the GQA dataset \cite{gqa} for visual question answering with an action space containing several actions for image understanding (e.g. localizing concepts in the image, counting objects, etc.) and image cropping. \tfontintro{\tfont{\initimgedit}} \cite{visualprogramming} requires an agent to edit images using a set of modules for image understanding and editing (e.g. face detection, segmentation, blurring). \tfontintro{\tfont{\initgaia}} \cite{gaia} requires question answering over various contexts (text, images, spreadsheets) using various GPT-4 plugins. The tasks were designed to be easy to solve for non-experts, but difficult for state-of-the-art LLM-based agents. \tfontintro{\tfont{\initmandm}} \cite{mms} requires question answering over text, image, and audio data, as well as image editing. The action space comprises tools for data understanding and editing. 

\tfontintro{\tfont{\inittoolbench}} \cite{toolbench} gives agents access to a large set of live REST API endpoints from the RapidAPI Hub\footnote{\url{https://rapidapi.com/hub}} covering different domains (e.g. Finance, Movies, Jobs, etc). %Some tasks require multiple actions to be executed, but usually there is no dependence between actions. 
\tfontintro{\tfont{\initrestbench}} \cite{restbench} provides agents with access to REST API endpoints for the Spotify music player for retrieving metadata and recommendations, creating and managing playlists, etc.  \tfontintro{\tfont{\inittoolalpaca}} \cite{toolalpaca} gives agents access to 11 real-world APIs providing information about holiday, transportation, entertainment, weather, etc.

%\begin{itemize}
   % \item Agents are required to solve a task by operating external pieces of software via Application Programming Interfaces (API)s, also referred to as tools\footnote{We consider APIs a complex class of tools}, 
    %\item APIs can be called in isolation, e.g. by indicating API name and arguments, or embedded in code 
   % \item Observations usually correspond to the APIs return values in case execution succeeds. In case of execution failure, information about error causes can be encoded into the error messages of exceptions thrown by the APIs directly (e.g. incorrect date formatting)
    %\item APIs vs tools: APIs are usually more complex than simple tools like a calculator, often domain specific
     %\item actions change world states or provide information
      %\item Recent tasks come with very large action spaces based on large amounts of APIs, and understanding the complex functionality of APIs poses one of the main challenges (select which API to use and how to call them) 
      %\item Recent agent methods resort to first retrieving a set of candidate APIs useful for solving a problem from the set of available APIs \mareike{CITE}, and equipping the agent with the ability to retrieve API documentation in order to decide on subsequent steps \mareike{CITE}
%\end{itemize}

\subsubsection{Interaction with GUIs}\label{sec:gui}
Instead of using predefined tools, digital assistants can also directly interact with graphical user interfaces (GUI) designed to be used by (non-expert) users, including websites, or GUIs of mobile phone and desktop applications. Action spaces comprise coordinate or id-based mouse and keyboard actions like \afont{click}, \afont{type}, or actions for interacting with mobile screens like \afont{press}, or \afont{swipe}. Observations correspond to representations of the current state of the GUI, e.g. in the form of screenshots, HTML, accessibility trees, often augmented with additional annotations like bounding boxes for indicating interactive elements in a screenshot \cite{visualwebarena}, or semantic labels for specific icons \cite{aitw}. 

\tfontintro{\tfont{\initmindtoweb}} \cite{mind2web} requires an agent to navigate websites based on their HTML representations, \tfontintro{\tfont{\initaitw}} \cite{aitw} focuses on navigation of mobile phone apps and websites based on annotated screenshots.  \tfontintro{\tfont{\initosworld}} \cite{osworld} requires agents to execute tasks in a computer environment, including both interaction with desktop applications and command line interfaces of different computer operating systems.

%\begin{itemize}
%\item Agents acting as digital assistants can also directly interact with graphical user interfaces (GUI) designed to be interacted with by (non-expert) users, including websites, or GUIs of mobile phone and desktop applications. 
%\item Action spaces comprise coordinate or id-based mouse and keyboard actions like \afont{click}, \afont{type}, etc. or actions for interacting with mobile screens like \afont{press}, or \afont{swipe}.
%\item Observations correspond to representations of the current state of the UI, e.g. in the form of screenshots, HTML, accessibility trees, often augmented with additional annotations like bounding boxes for indicating interactive elements in a screenshot \mareike{CITE}, or semantic labels for specific icons \mareike{CITE}. Choosing the best modality for representing observations is an active research question \mareike{CITE}, as well as how to shape the action space \mareike{CITE}
%    \item Main challenges comprise a large action space (small number of actions with large number of possible parameters, e.g. buttons to interact with), which is often tackled by pre-filtering elements to interact with \mareike{CITE}, and extracting relevant information from the observation/understanding the representation of the UI \cite{visualwebbench}, which is possibly the reason for a large number of tasks being released as trajectory-only  datasets, and agents are evaluated step-by-step.
%\end{itemize}

%\paragraph{Travel planning}
%\subsection{Question answering}\label{sec:qa}
%\paragraph{Mathematical reasoning}\label{math:qa}
\subsubsection{Interaction with Code Interpreters}\label{sec:programming}
Digital agents can also directly interact with code interpreters, e.g. to execute %SHORT SQL commands, 
or python programs. The action space is the set of all valid statements in the respective programming language, which can include external libraries corresponding to APIs, or directly enabling actions for GUI interaction. 

For example, \tfontintro{\tfont{\osworld}} enables an agent to navigate a GUI via a python interpreter and the python \texttt{PyAutoGUI} library\footnote{\url{https://pyautogui.readthedocs.io}}. Observations directly correspond to outputs of the interpreter. \tfontintro{\tfont{\initdatabase}} \cite{agentbench} tasks are based on a collection of existing datasets for database QA \cite{wikisql,wikitablequestions,sqa,hybridqa,fetaqa}, i.e. answering the question requires reading information from databases tables, and the action space corresponds to the set of valid SQL commands. \tfontintro{\tfont{\initappworld}} \cite{appworld} requires agents to use a set of day-to-day applications via APIs, and enables agents to directly interact with a python interpreter.

\section{Structuring the Task Landscape}\label{sec:tasks}
\begin{table*}[ht]
\resizebox{\textwidth}{!}{
\centering
\begin{tabular}{ llllllllll }\toprule

 &\textbf{Task  }  &\textbf{Environment }  & \textbf{Application}  &  \rotatebox[origin=c]{90}{\textbf{ Obs }}  & \rotatebox[origin=c]{90}{\textbf{ C}}   & \rotatebox[origin=c]{90}{\textbf{ IR}}   &  \rotatebox[origin=c]{90}{\textbf{ AS }}  &    &  \textbf{Evaluation } \\
 
\multirow{18}{*}{\rotatebox[origin=c]{90}{Goal: reach world state}} &\tfont{\blocksworld} & PDDL & N\&M& \CIRCLE & \checkcolor{\ding{51}} &  \crosscolor{\ding{55}} & F(a) & & GS\\

&\cellcolor{tablegrey}\tfont{\scienceworld}  & \cellcolor{tablegrey}Text world & \cellcolor{tablegrey}N\&M& \cellcolor{tablegrey}\LEFTcircle & \cellcolor{tablegrey}\checkcolor{\ding{51}} & \cellcolor{tablegrey}MA & \cellcolor{tablegrey}F(a) &\cellcolor{tablegrey}  & \cellcolor{tablegrey}Reward\\

&\tfont{\alfworld} & Text world& N\&M & \LEFTcircle & \checkcolor{\ding{51}} &  \crosscolor{\ding{55}} & F(a) &  & GS\\

&\cellcolor{tablegrey}\tfont{\jericho} & \cellcolor{tablegrey}Text world& \cellcolor{tablegrey}N\&M & \cellcolor{tablegrey}\LEFTcircle & \cellcolor{tablegrey}\checkcolor{\ding{51}} & \cellcolor{tablegrey}MA & \cellcolor{tablegrey}NL &\cellcolor{tablegrey}  & \cellcolor{tablegrey}GS\\

&\tfont{\gridlu} & \twod world& N\&M & \CIRCLE & \checkcolor{\ding{51}} &  \crosscolor{\ding{55}} & F(a) &  & GS\\

&\cellcolor{tablegrey}\tfont{\minigrid} & \cellcolor{tablegrey}\twod world& \cellcolor{tablegrey}N\&M & \cellcolor{tablegrey}\LEFTcircle & \cellcolor{tablegrey}\checkcolor{\ding{51}} &  \cellcolor{tablegrey}\crosscolor{\ding{55}} & \cellcolor{tablegrey}F(a) &\cellcolor{tablegrey}  & \cellcolor{tablegrey}Reward \\

&\tfont{\alfred} & \threed world& N\&M & \LEFTcircle & \checkcolor{\ding{51}} &  \crosscolor{\ding{55}} & F(a) &  &  Partial GS\\

&\cellcolor{tablegrey}\tfont{\minedojo}  & \cellcolor{tablegrey}Minecraft & \cellcolor{tablegrey}N\&M& \cellcolor{tablegrey}\LEFTcircle & \cellcolor{tablegrey}\checkcolor{\ding{51}} & \cellcolor{tablegrey}MO & \cellcolor{tablegrey}F(a) &\cellcolor{tablegrey}   & \cellcolor{tablegrey}Model (fine-tuned)\\

&\tfont{\minerl} & Minecraft& N\&M & \LEFTcircle & \checkcolor{\ding{51}} &  \crosscolor{\ding{55}}  & F(a) &  & Human (comparison)\\

&\cellcolor{tablegrey}\tfont{\imgedit}  & \cellcolor{tablegrey}Tools/APIs&\cellcolor{tablegrey}Img. editing & \cellcolor{tablegrey}\LEFTcircle & \cellcolor{tablegrey}\checkcolor{\ding{51}} &  \cellcolor{tablegrey}\crosscolor{\ding{55}} &\cellcolor{tablegrey} F(a) &\cellcolor{tablegrey}  & \cellcolor{tablegrey}Human (correctness)\\

&\tfont{\mandm} & Tools/APIs&\makecell[l]{Img. editing\\QA (Img.)} & \LEFTcircle & \checkcolor{\ding{51}} &  \crosscolor{\ding{55}} & F(a) &  & unordered RT (F)\\

&\cellcolor{tablegrey}\tfont{\restbench}  &\cellcolor{tablegrey}Tools/APIs&\cellcolor{tablegrey}Digital assistance & \cellcolor{tablegrey}\LEFTcircle & \cellcolor{tablegrey}\checkcolor{\ding{51}} &  \cellcolor{tablegrey}\crosscolor{\ding{55}} &\cellcolor{tablegrey} F(a) &\cellcolor{tablegrey}  & \cellcolor{tablegrey}\makecell[l]{RT (subsequence) \\ \cellcolor{tablegrey}Human (correctness)}\\

&\tfont{\appworld} &Tools/APIs (via code)&Digital assistance & \LEFTcircle & \checkcolor{\ding{51}} &  \crosscolor{\ding{55}} & PY &  & GS (constraints)\\

&\cellcolor{tablegrey}\tfont{\aitw} & \cellcolor{tablegrey}GUI& \cellcolor{tablegrey}Digital assistance & \cellcolor{tablegrey}\LEFTcircle &  \cellcolor{tablegrey}\checkcolor{\ding{51}} &  \cellcolor{tablegrey}\crosscolor{\ding{55}} &\cellcolor{tablegrey} F(a) & \cellcolor{tablegrey} & \cellcolor{tablegrey}RT (step-by-step)\\

&\tfont{\mindtoweb} & GUI& Digital assistance& \LEFTcircle & \checkcolor{\ding{51}} &  \crosscolor{\ding{55}} & F(a) &  & RT (step-by-step)\\

& \cellcolor{tablegrey}\tfont{\osworld} & \cellcolor{tablegrey}GUI (via code)& \cellcolor{tablegrey}Digital assistance  & \cellcolor{tablegrey}\LEFTcircle & \cellcolor{tablegrey}\checkcolor{\ding{51}} &  \cellcolor{tablegrey}\crosscolor{\ding{55}} & \cellcolor{tablegrey}PY &\cellcolor{tablegrey}  & \cellcolor{tablegrey}GS\\

\addlinespace[20pt]

\multirow{9}{*}{\rotatebox[origin=c]{90}{Goal: answer a question}} &\cellcolor{tablegrey}\tfont{\embodiedqa}  &\cellcolor{tablegrey}\threed world& \cellcolor{tablegrey}QA (Physical contexts)& \cellcolor{tablegrey}\LEFTcircle & \cellcolor{tablegrey}\checkcolor{\ding{51}} &  \cellcolor{tablegrey}\crosscolor{\ding{55}} & \cellcolor{tablegrey}F(a) & \cellcolor{tablegrey} & \cellcolor{tablegrey}RA\\

  &\tfont{\hotpot}&Tools/APIs& QA (Wiki facts)& \LEFTcircle & \crosscolor{\ding{55}} &  \crosscolor{\ding{55}} & F(a) &  & RA\\

&\cellcolor{tablegrey}\tfont{\gaia} &\cellcolor{tablegrey}Tools/APIs& \cellcolor{tablegrey}QA (Multimodal contexts) &\cellcolor{tablegrey} \LEFTcircle & \cellcolor{tablegrey}\crosscolor{\ding{55}} &  \cellcolor{tablegrey}\crosscolor{\ding{55}} & \cellcolor{tablegrey}F(a) &\cellcolor{tablegrey}  & \cellcolor{tablegrey}RA\\

&\tfont{\gsmeightk}  &Tools/APIs& QA (Math contexts) & \LEFTcircle & \crosscolor{\ding{55}} &  \crosscolor{\ding{55}} & F(a) &  & RA\\

&\cellcolor{tablegrey}\tfont{\database} &\cellcolor{tablegrey}Tools/APIs& \cellcolor{tablegrey}QA (KG) & \cellcolor{tablegrey}\LEFTcircle & \cellcolor{tablegrey}\crosscolor{\ding{55}} & \cellcolor{tablegrey} \crosscolor{\ding{55}} & \cellcolor{tablegrey}F(a) &\cellcolor{tablegrey}  & \cellcolor{tablegrey}RA\\

&\tfont{\gqa}  &Tools/APIs& QA (Img.) & \LEFTcircle & \crosscolor{\ding{55}} &  \crosscolor{\ding{55}} & F(a) &  & RA\\

&\cellcolor{tablegrey}\tfont{\toolbench}  &\cellcolor{tablegrey}Tools/APIs& \cellcolor{tablegrey}Digital assistance & \cellcolor{tablegrey}\LEFTcircle & \cellcolor{tablegrey}\crosscolor{\ding{55}} &  \cellcolor{tablegrey}\crosscolor{\ding{55}} &\cellcolor{tablegrey} F(a) & \cellcolor{tablegrey} & \cellcolor{tablegrey}\makecell[l]{Model (ICL;correctness) \\ \cellcolor{tablegrey}Model (ICL;comparison)}\\

&\tfont{\toolalpaca} &Tools/APIs& Digital assistance & \LEFTcircle & \crosscolor{\ding{55}} &  \crosscolor{\ding{55}} & F(a) &  & Model (ICL;correctness)\\

\bottomrule
\end{tabular}
}
\caption{ Overview over agent tasks which require reaching a specific world state (upper part), or answering a question (lower part). N\&M = navigation and object manipulation, for QA applications, we indicate relevant contexts \cite{rogers2023qa} in brackets. \textbf{Obs} = Observability of state ($\CIRCLE$ = full, $\LEFTcircle$ = partial). \textbf{C} = Agent can change world states. \textbf{IR} = Intermediate rewards (MA = designed manually, MO = model-based). \textbf{AS} = action space (F(a) = parametric, NL = natural language, PY = python code). Evaluation: GS = goal state, RA = reference answer, RT = reference trajectory, ICL = in-context learning. For \tfont{\restbench}, we display the \emph{Spotify} part of their dataset.}
\label{ t:detailed }
\end{table*}
In the following, we provide a detailed description and comparison  of the different characteristics of complex tasks for evaluating agent performance, pointing out how these characteristic contribute to challenges for agents. \cref{ t:detailed } presents the tasks introduced in the previous section, illustrating differences in task components. %The complete overview over the tasks included in our survey can be found in Appendix and on our project website\footnote{\url{https://docs.google.com/spreadsheets/d/19kW4PHiXHt9MenT0y_xF676TB7v2vszgAj_DaHBoj6I/edit\#gid=1719500727}}. 

%We introduce some central concepts relevant to interactive goal-directed agent tasks, we briefly introduce the most common paradigms for implementing 

%\begin{itemize}
    %\item Task: In general, the task is to come up with a sequence of actions to achieve a given goal (as a type of NLP task). We consider different instantiations of this task (with significant difference in action spaces, state spaces, observation spaces, e.g. text games vs Minecraft) as different tasks.
    %\item We consider as \problem: A combination of goal, environment configuration (including initial world state) given to the agent \mareike{RL calls this 'episode'?}
%\end{itemize}

\subsection{Task Formalization}\label{sec:formalization}

An agent's objective is to come up with a course of action in order to achieve a goal in a given environment. In order to do so, it interacts with the environment\footnote{Following \citet{sutton2018reinforcement}, we consider the \textit{environment} to comprise anything outside the agent.} in (discrete) time steps by taking actions, and observing the actions' effect on the environment. 
We formally define a task instance as a Partially Observable Markov Decision Process (POMDP) $\langle S,A, T, O, \Omega \rangle$, augmented with an initial state $S_0$ and a goal specification $G$. $A$ is a set of actions, $S$ is a set of states, and $T$ a state transition function $T :  S \times A \rightarrow S$ specifying a state transition from $s$ to $s'$ in case action $a$ is taken. We define the set of \textit{admissible actions} $A(s)$ as actions available in a given state, i.e. that will lead to a state transition if executed in $s$. 

States correspond to situations the agent can modify via its actions. %: In some cases, they can correspond to \textit{\worldstates} which the agent can manipulate, e.g. when the task requires to navigate a  grid and manipulate objects. In other cases, they correspond to \textit{\kstates}, which the agent can extend by retrieving or combining pieces of information, e.g. when the tasks requires retrieving facts from a knowledge base and drawing inferences based on these facts. 
The agent receives information about the current state via observations $o$ from the set of observations $O$ according to the observation function $\Omega: S \times A \rightarrow  O$. $G$ is a specification of the goal, which varies by goal type and how directly it expresses the goal conditions (see \cref{sec:goals}). %$R$ is a reward function which assigns a scalar reward to each combination of state and action as $R : S \times A \rightarrow [0,1]$. %Most of the tasks in this survey assign a positive reward only upon successful task completion, i.e. $R(s, a) = 1$ iff the goal conditions are satisfied and 0 otherwise. Some tasks additionally provide intermediate rewards, i.e. positive rewards are assigned to indicate progress towards meeting the goal conditions. 
   
Given $G$ and an observation of $S_0$, the agent's objective is to come up with a sequence of actions from $A$ to complete the goal specified by $G$ by interacting with the environment: the agent taking an action affects the environment, which in turn emits an observation affecting the agent's next action. %At each time step, the agent takes an action informed by various pieces of information: information about the current state $s$, the transition function $T$ (often referred to as the world model), and previous interactions. Agent architectures greatly differ in how explicitly they make each of these pieces of information.

%\section{Task Dimensions \mareike{Task Components}}\label{sec:dimensions}

\subsection{Goals}\label{sec:goals}
\paragraph{Goal specification}
The goal specification $G$ conveys information about the conditions in which the task is considered completed. %, i.e. $S_G$ or $A_G$. 
It is a NL expression in form of an instruction or question. Specifications vary in how explicitly they specify the goal conditions, ranging from direct NL translations of goal states (e.g. \exfont{Green triangle west of a red circle.} in \tfont{\gridlu})
to less direct specifications expressing the constraints a goal state needs to satisfy (e.g. \exfont{Play my Spotify playlist with enough songs for the entire workout today.} in \tfont{\appworld}). 
Less direct goal specifications contribute to task difficulty, as the agent cannot directly work towards reaching the goal state, but first needs to acquire more information on what a valid goal state looks like. This usually means that the task can be broken into sub-tasks, favoring agents with explicit mechanisms for task decomposition \cite{khot2023decomposed,wang2023plan,adapt,rada}. %For some tasks, goal specifications also contain explicit hints on how to establish the goal conditions (e.g. the addition \exfont{My workout plan is in SimpleNote} in the \tfont{\appworld} task example above.).

\paragraph{Goal type}
One fundamental difference between tasks is what type of goal needs to be completed, which dictates how we can meaningfully evaluate if a task was completed. We distinguish two goal types (upper and bottom part in \cref{ t:detailed }):
    \begin{enumerate}
        \item The goal is to reach a specific world state. The goal specification $G$ can be mapped to a set of goal states $S_G \subseteq S$, and the goal is achieved if the current world state $s \in S_G$. %Natural language goal specifications for these types of tasks are usually expressed as instructions.
        \item The goal is to answer a question. The goal specification $G$ can be mapped to a subset of actions $A_G \subseteq A$, and the goal is achieved if the agent decides to take action $a \in A_G$, usually to submit a final answer to the question.
    \end{enumerate}

\paragraph{Stopping criteria}
In most environments, agents need to perform a dedicated final action indicating that they established the goal conditions, e.g. a \afont{stop} action indicating a goal state was reached (or an \afont{answer} action providing a final answer). Some environments %do not require such explicit indication and 
recognize a task as completed whenever the agent reaches a goal state. This simplifies the task, as it does not require the agent to recognize goal completion.

%\paragraph{Goal conditions}
%Most tasks included in our survey specify goal conditions which can objectively be assessed as satisfied or violated, e.g. by checking if the agent's end state satisfies specific constraints, or by comparing the agent's answer with a reference answer. For other tasks, completion cannot be objectively evaluated. This applies to creative tasks like \exfont{build an epic modern house with two floors and a swimming pool} (\tfont{MineDojo}), and subjective tasks like \exfont{Download a funny joke from platform X.} (\tfont{ToolBench}). Evaluation of task completion for such tasks poses a challenge, which we will further discuss in \cref{sec:evaluation}. \mareike{should this be moved to evaluation section? It has implications for evaluation, but not for the agent?}
       
%\mareike{The tasks included in this survey specify goal conditions via hard constraints, i.e. all constraints need to be satisfied in order for the goal conditions to be met. Soft constraints, on the other hand, are constraints which should ideally be satisfied but can be violated to a certain extent if it benefits the overall result. Many real-world problems can be modelled via soft constraints, e.g. as a reflection of user preferences EXAMPLE. We expect that in the future, more tasks with soft constraints will be introduced.}

%\item Some tasks have multiple possible goal states, or multiple possible answers \mareike{But this depends on evaluation, so address this in Evaluation section?}

\subsection{World and Knowledge States}\label{sec:states}
Many tasks require agents to manipulate \worldstates, i.e. change a situation such that it satisfies the goal conditions (e.g. manipulate objects, modify database states). Other tasks require the agent to retrieve %SHORT , extract, 
or transform information about a specific situation or context without manipulating the \worldstate (e.g. multi-hop QA tasks or knowledge-base QA tasks)\footnote{This fundamental difference in tasks has been noted widely and referred to as contrast between \textit{embodied reasoning vs. language reasoning} \cite{law}, \textit{environment-in-the-loop vs. tool-in-the-loop} \cite{interactivenlp}, \textit{decision-making vs. reasoning} \cite{mint}}. In this survey, we adopt a wide definition of the concept of a state: States can correspond to \emph{\worldstates}, e.g. a \twod grid, or a \threed simulation of a kitchen. States can also correspond to \emph{\kstates}, i.e. a collection of information, e.g. related to the real world, to a document, or to entries in a knowledge graph.

%\item A state $s \in S$ has two components: $x$ is the current \textit{\worldstate}, i.e. a set of statements describing a situation. $k$ is the agent's \textit{knowledge state}, which is the subset of statements in $x$ known\mareike{/available} to the agent. Tasks differ in if they enable the agent to manipulate the \worldstate, which depends on the actions the agent can choose.

\paragraph{State changes}
If the agent can modify the \worldstate, it can transition to situations where actions required for task completion are inadmissible, and needs to revert previous actions before being able to take the necessary step (e.g. locking a door requires unlocking it before opening it to step on the other side). In a worst case, executing irreversible \worldstate changes can lead to dead-end states, i.e. situations in which tasks become unsolvable, e.g. by destroying gold with a laser weapon in \tfont{\goldminer}, or deleting DB entries such as amazon orders in \tfont{\appworld}. %\mareike{Need for world state tracking?}
If states correspond to \kstates, %SHORT the agent's 
actions serve to acquire or transform information, which %SHORT usually 
does not render any follow-up actions inadmissible.

Some environments enable environment-based state changes, i.e. world state changes independent of agent actions (e.g. a bird hatching from an egg in \tfont{\scienceworld}, or day-to-night progression in Minecraft environments), which adds to the complexity of the world model the agent needs to maintain in order to master the 
environment.

\paragraph{Observability of \worldstates}
A state $s$ is \textit{fully observable} if the agent observes perfect information about $s$, in which case $\Omega(s,a) = s$, and \textit{partially observable} otherwise. Partially observable scenarios require the agent to perform actions in order to gather more information about the current state, before working towards the goal. %\mareike{sometimes referred to as exploration, but in RL exploration means: Agent takes an action that is sub-optimal according to the current policy (does not have the highest Q-value), which often increases long-term rewards.}. 
Full observability in realistic tasks involving real-world problems is rare (e.g. board games without stochasticity such as chess), and usually only occurs in synthetic tasks with small/low-dimensional \worldstates, e.g. placing objects in a 9x9 grid as in \tfont{\gridlu}, or stacking a limited amount of blocks as in \tfont{\blocksworld}.

%\mareike{Observability of states that are \kstates?} 
%\begin{itemize}
%\item History is more important because of failed attempts?
%    \item Possibly fully observable via information gathering actions (e.g. \alfworld), infeasible to fully observe (e.g. QA), impossible to fully observe (by design, e.g. because sensing does not reflect the real world state)
%\end{itemize}

\subsection{Actions}\label{sec:actions}
%Action spaces differ in size and nature, and the current trend is to introduce tasks with larger action spaces. 
The action spaces of most of the tasks included in this survey can naturally be modelled as \emph{parameterized action spaces}, i.e. combinations of discrete actions with parameters \cite{masson2016reinforcement}. The set of possible parameters can be continuous, discrete and possible to enumerate\footnote{Even if it is theoretically possible to enumerate all possible parameters, many task environments do not readily provide support to enumerate them (e.g. \tfont{\alfworld}).}
(e.g. all objects to interact with in a given room), or too large to explicitly enumerate (e.g. all NL queries as a parameter for a search engine). Some tasks come with very high-dimensional discrete action spaces, e.g. the set of all valid python programs, or the set of all NL sentences, which combinatorially grows with the vocabulary size
. 
Whereas small action spaces allow the explicit enumeration of all possible actions, which can then be scored \cite{tan2024true}, or explicitly listed in the prompt of an LLM-based agent \cite{react}, this is infeasible for very large action spaces. Recent work resorts to filtering action spaces to only keep top-k candidate actions at each time step \cite{mind2web}, or restrict the action space permanently given the goal specification in the initial state \cite{gorilla,toolbench}.

\paragraph{Action preconditions and effects}

To decide on the right actions for solving a task, the agent must have some knowledge about the transition function $T$, in particular about \textit{action preconditions},  i.e. the constraints $s$ must satisfy such that $T(a, s) = s'$, and the \textit{action effects} on the state. The agent can learn about the transition function through interaction, i.e. by trial-and-error. If knowledge about the transition function can be appropriately formalized, it can be build into the agent directly%without the agent having to interact with the environment to observe action effects
. Some tasks provide full specifications of the transition function, e.g. PDDL planning problems such as \tfont{\blocksworld}. Here, action preconditions and effects are fully described. In combination with fully observable world states, agents with full knowledge about the transition function can reach goal states via search, without requiring any interaction with the environment \cite{helmert2006fast}.

\paragraph{Planning and execution time effects}
Many tasks with parameterized action spaces come with \textit{partial} descriptions of action preconditions and effects. These convey a \textit{planning time effect}  \cite{bacchus1998modeling}: the part of the action effect known at planning time (as opposed to the \textit{execution time effect}, which is the effect observable at execution time). For example, the action \afont{get\_password(user)} has the planning time effect of the agent receiving the password of a user, the execution time effect is that the action will deliver the value of this password (e.g. \afont{password='1234'}). The planning time effect can be thought of as a docstring comment for a function, giving abstract information about the functions workings, but not about concrete return values.
Parameterized action spaces often come with descriptions of planning time effects, but not execution time effects of all possible combinations of action and parameters. As building information about all action planning time effects into the agent directly is infeasible for large action spaces, some tasks (e.g. \tfont{\appworld}) provide functionality for retrieving descriptions of planning time effects (e.g. in the form of API documentation) on demand, i.e. as an additional action the agent can decide to perform. %SHORT, providing the agent with partial information about the transition function.

\subsection{Observations}\label{sec:observations}
\begin{table*}[t]
\resizebox{\textwidth}{!}{
    \centering
    \begin{tabular}{lllp{6.6cm}l}
    %\textbf{Task} & \textbf{Example goal specification} & \textbf{Comment} \\
    %\toprule
              %&\multicolumn{4}{c}{\textbf{Observations upon admissible actions}}  \\
              &\makecell[c]{\textbf{Task}} & \makecell[c]{\textbf{Action}} & \makecell[c]{\textbf{Observation}} & \makecell[c]{\textbf{Comment}} \\
      %& \tfont{\minigrid} & &  \makecell[c]{\adjustbox{valign=c}{\includegraphics[width=1.5cm, height=1.5cm]{images/minigrid_newstate.png}}} & full updated world state \\
       %\addlinespace[5pt]
       
        \multirow{6}{*}{\rotatebox[origin=c]{90}{$a \in A(s)$}} &\cellcolor{tablegrey}\tfont{\blocksworldshort} &\cellcolor{tablegrey}\texttt{unstack block\_1 from block\_2} & \cellcolor{tablegrey}\exfonttable{Block\_1 is now unstacked from block\_2.} & \cellcolor{tablegrey}immediate action effect \\
       
       &  \tfont{\alfworld} & \texttt{go to shelf 6} &  \makecell[l]{\exfonttable{You arrive at loc 4. \underline{On the shelf 6,}}\\\exfonttable{\underline{you see a vase 2.}}} & \makecell[l]{immediate action effect +\\\underline{prev. unobserved information}} \\
       
        &\cellcolor{tablegrey}\tfont{\toolbench} & \cellcolor{tablegrey}\makecell[l]{\texttt{Func: get\_personal\_details} \\\texttt{Args: \{'name': 'Bob Smith'\} }} & \cellcolor{tablegrey}\makecell[l]{\{\exfont{'age': 28,} \\ \exfont{\ \ 'recent movies':} [ \exfont{'Spider-Man: Across}\\\exfont{\ \ \ the Spider-Verse'}, $\cdots$ ]\}} & \cellcolor{tablegrey}function output \\
       %\multicolumn{4}{c}{\textbf{Observations upon inadmissible actions}}  \\
       \addlinespace[18pt]

       \multirow{6}{*}{\rotatebox[origin=c]{90}{$a \notin A(s)$}} & \cellcolor{tablegrey}\tfont{\alfworld} & \cellcolor{tablegrey}\texttt{sit down}& \cellcolor{tablegrey}\exfonttable{Nothing happens.} & \cellcolor{tablegrey}no additional feedback \\
       
        &\tfont{\blocksworldshort} &\texttt{stack object\_3 object\_1} & \makecell[l]{\exfonttable{I cannot stack object\_3 on top of} \\
        \exfonttable{object\_1 because \underline{object\_1 is not clear.}}} & \makecell[l]{specification of \\ \underline{violated action preconditions}} \\ 
        
        &\cellcolor{tablegrey}\tfont{\appworld} &\cellcolor{tablegrey}\texttt{search\_contacts\_for\_phone()}& \cellcolor{tablegrey}\makecell[l]{\{\exfonttable{'error': 'You are either not authorized} \\ \exfont{\ \ to access this API endpoint or your acc.} \\ \exfont{\ \ token is missing, invalid or expired.'} \}}& \cellcolor{tablegrey}customized error messages \\ 
      %&\tfont{\appworld} &  & \exfonttable{} & Python interpreter stack trace \\ 
    \end{tabular}}
    \caption{Examples for differences in observations upon execution of admissible actions (upper part), and inadmissible actions (lower part). \tfont{\blocksworld} is shortened to \tfont{\blocksworldshort} for space reasons.}
    \label{tab:observations}
\end{table*}
We consider as \emph{observation} any information the environment exposes to the agent as a result of action execution. Tasks differ in how much and what type of information an observation conveys, which usually includes the immediate effects of action execution, and possibly feedback in case of execution failures, or feedback on progress towards task completion (intermediate rewards). Examples %SHORT of differences in the information content of observations 
can be found in \cref{tab:observations}.

\paragraph{Observing action execution effects}
For world state transitions, observations convey information about how action execution impacts the \worldstate, e.g. by exposing (the observable part of) 
the new \worldstate. Often, %SHORT in particular when states are high-dimensional, 
the observation exposes information on what changed due to the state transition, i.e. the difference between source and target state, requiring the agent to derive the new \worldstate based on this information. For information retrieval/transformation actions, observations directly correspond to function outputs, e.g. the output of a function looking up information in a knowledge base.

\paragraph{Observing feedback on inadmissible actions}
In some setups, agents might decide to take \textit{inadmissible} actions $a \notin \mathrm{A}(s)$, i.e. actions which cannot be executed in a given state, for example when sequence-based agent architectures generate actions with an LLM \cite{decisiontransformer} rather than scoring a pre-defined set of admissible actions. An action is inadmissible either because the state does not satisfy the action's preconditions (e.g. trying to walk through a closed door), or because the action is \textit{invalid} ($a \notin \mathrm{A}$), e.g. when combining a parameterized action with an incorrect parameter, such as passing a parameter of incorrect type to a python method. Tasks differ in how informative observations are in such cases, e.g. only providing the information that an action had no effect, or detailed feedback on reasons for execution failure%SHORT(see \cref{tab:observations})
. 

 \paragraph{Observing intermediate rewards}
 For some tasks, observations additionally provide intermediate rewards, i.e. per-step feedback on if and how much an action contributed to task completion. This signal can either be manually crafted, or automatically calculated by a model: \tfont{\scienceworld} manually annotates sub-goals for each goal specification and provides intermediate scalar rewards corresponding to fractions of completed sub-goals, \tfont{\agentboard} \cite{agentboard} augments several existing environments with intermediate rewards in the same way. \tfont{\llfbench} \cite{llfbench} extends existing environments with different types of verbalized feedback, including a verbalization of a scalar reward signal. \tfont{\minedojo} provides intermediate rewards via a reward model trained on pairs of goal specification and sequences of observations in the form of Youtube videos. \tfont{\mint} \cite{mint} extends existing environments with verbalized intermediate reward feedback generated by an LLM. Whereas these reward models can be considered part of the respective environments (as they were released along with the tasks), many recent agent architectures implement their own components for generating intermediate feedback, which are considered part of the agent rather than the environment (e.g. \citealp{reflexion}).

\paragraph{Modality of the observation space}
Depending on the nature of the task, observations are represented in various modalities, which informs what models can be used for processing the information (e.g. LLMs vs multi-modal models). For example, \twod or \threed states are usually observed via visual information, but could also be observed via NL descriptions (e.g. \tfont{\alfworld} as a verbalized version of \threed \tfont{\alfred}). Observations can also come as structured text, e.g. outputs of APIs often come as json dictionaries, or HTML accessibility trees for web navigation tasks. Depending on the nature of the observation, agents might struggle to extract relevant information from the observation object. For example, observations of webpages via HTML have been shown to be particularly difficult for LLM-based agents to process \cite{visualwebbench}. %SHORT So far, it is unclear which observation modality makes information most accessible to an agent, and 
Identifying the best modality for representing observations is an active research question \cite{webarena, visualwebarena, osworld}, and some tasks (e.g. \tfont{\osworld}) provide a range of different observation representations. For example, the GUI interaction tasks in \tfont{\osworld} come with a range of different representations for representing GUI states, including html-based accessibility trees, or screenshots with or without set-of-marks annotations \cite{som} indicating relevant elements to interact with. %\textcolor{blue}{Choosing the best modality for representing observations is an active research question \mareike{CITE}, as well as how to shape the action space \mareike{CITE}. The main challenges comprise a large action space (small number of actions with large number of possible parameters, e.g. buttons to interact with), which is often tackled by pre-filtering elements to interact with \mareike{CITE}, and extracting relevant information from the observation/understanding the representation of the UI \cite{visualwebbench}, which is possibly the reason for a large number of tasks being released as trajectory-only datasets, and agents are evaluated step-by-step.}

 %\begin{itemize}  
 %\item \tfont{\llfbench} extends existing environments with different types of verbalized feedback: verbalization of scalar reward signal
 %\item \tfont{\mint} verbalized intermediate/reward feedback generated by an LLM
%\item Observations are objects that serve to provide agents with information about effects of action execution. Tasks differ in how much information an observation provides. Generally, observations convey the effect of action execution. For world state transitions, they convey information about how action execution impacts the \worldstate, e.g. by exposing (the observable part) of the new \worldstate. Often, in particular when states are high-dimensional, the observation exposes information on what changed with the state transition, i.e. the difference between source and target state, requiring the agent to derive the new \worldstate based on the diff information. For actions for information retrieval/transformation, observations directly correspond to function outputs, e.g. the output of a function looking up information in a knowledge base. Some task environments additionally provide feedback on reasons for failed action execution, or global feedback on if and how much an action contributed to progress towards task completion (intermediate rewards). Examples are shown in \cref{tab:goalspecifications}.

%\end{itemize}
%\begin{itemize}

%\end{itemize}

\subsection{Task Evaluation}\label{sec:evaluation}
Task evaluation serves to determine if an agent successfully completed a given task, i.e. established goal conditions. Most tasks included in our survey specify goal conditions which can objectively be assessed as satisfied or violated, e.g. by checking if the agent's end state satisfies specific constraints, or by comparing the agent's answer with a reference answer. For other tasks, completion cannot be objectively evaluated, e.g. for creative tasks like \exfont{build an epic modern house with two floors and a swimming pool} (\tfont{\minedojo}), and subjective tasks like \exfont{Download a funny joke from platform X} (\tfont{\toolbench}). A range of different evaluation methods have been suggested for evaluating task completion.

%%Evaluation of task completion for such tasks poses a challenge, which we will further discuss in \cref{sec:evaluation}. %\mareike{should this be moved to evaluation section? It has implications for evaluation, but not for the agent?}
\paragraph{Reference-based evaluation of final answers}
For most QA tasks, a predicted answer is compared to a reference answer, via
exact match, fuzzy match, or based on the reference answers rank in a predicted ranking \cite{embodiedqa}. Several scenarios make comparison to static ground truth answers infeasible: creative or subjective tasks as illustrated above, and non-controllable data sources that might change over time (e.g. web-based APIs as in \tfont{\toolbench}). %SHORT In such scenarios, other evaluation options are reference-based evaluation of action sequences, or reference-free evaluation.

\paragraph{Reference-based evaluation of final states}
Tasks specifying goal states can be evaluated based on the final state the agent transitioned to, usually by evaluating if constraints of a reference goal state are satisfied. Some tasks evaluate partial goal completion, based on the final state's distance from the goal state, or the amount of constraints the final state satisfies. 

\paragraph{Reference-based evaluation of action sequences}
Here, predicted action sequences are compared to human-annotated reference sequences, either based on exact match or fuzzy match (checking if the predicted trajectory is a subsequence of the reference trajectory). \citet{mms} (\tfont{\mandm}) calculate precision and recall of predicted actions with respect to a set-based representation of the reference sequence.\citet{toolalpaca} (\tfont{\toolalpaca}) use GPT4 \cite{gpt4} to score predicted trajectories %and predicted final answers 
with respect to human reference trajectories. Reference-based evaluation of action sequences is a conservative metric, as it assumes a single correct action sequence for satisfying goal conditions, whereas most tasks can be solved in multiple ways.

\paragraph{Reference-free evaluation}
The evaluation methods described above require comparison to a reference answer, trajectory, or goal state. Instead, reference-free approaches evaluate generations %SHORTed answers, action sequences or final states by a human or an LLM. 
by a human or an LLM. The former ask humans to judge the correctness of final answers and/or predicted action sequences.\footnote{We found that annotation guidelines and precise descriptions of what outputs humans are asked to judge 
are often underspecified, likely leading to irreproducible and inconsistent
evaluations.} \citet{basalt2022} (\tfont{\minerl}) ask human evaluators to compare two trajectories predicted by different systems, and train a classifier on the resulting dataset to automatically compare predicted trajectories. Several works rely on LLMs in zero or few-shot fashion, prompting the LLM to evaluate the quality of predicted answers, to judge if a task was successfully solved given an instruction and a trajectory, or to compare action sequences generated by two different agents \citet{toolbench}. If LLMs provide a reliable means of evaluation is currently an open question, as recent work showed large variance across tasks and models in how well LLM judgments correlate with human judgments 
\cite{llmeval}.

\subsection{General Properties of the Environment}\label{sec:envproperties}
\paragraph{Indicators of task difficulty}
The community aims to introduce more and more challenging tasks, raising the question what makes a task difficult. \citet{osworld} define task difficulty according to human completion time. Several datasets provide annotations of difficulty levels, either by the authors \cite{appworld} or an LLM \cite{infiagent}, which express a subjective inherent difficulty of a task mainly useful for model analysis.

An objective measure of task difficulty could be helpful for both task creation and gaining insights from analysing agent performance. Several objective indicators have been discussed, most commonly the length of a gold trajectory for solving the task, i.e. the minimum amount of actions required to establish goal conditions \cite{codeact,taskbench,gaia}, empirically shown to impact agent performance \cite{autoplanbench, appworld}. Other indicators include the size of the action space \cite{toolbench}, the number of \emph{different} actions required to solve a task \cite{appworld, gaia}, and the number of objects to interact with %SHORT, which impacts the size of both action and state spaces 
\cite{textworld}.

\paragraph{Domain specificity}
As domain specificity, we consider the nature of the environment impacting the (mis-)alignment between task-relevant knowledge and knowledge in LLM pre-training data. Some tasks come with environments based on simulations of our real world (e.g. \tfont{\alfred} as a realistic \threed simulation of a household), and relevant task knowledge, e.g. on action preconditions and effects, state transitions, or optimal policies, corresponds to common knowledge. For other tasks, relevant task knowledge is not common but rather domain-specific, e.g. how to utilize operating systems as required in \tfont{\osworld}. We expect such tasks to be more challenging for LLM-based agents, as specialized knowledge might rarely occur in the pre-training data and hence not be stored in the models' parametric knowledge \cite{mallen-etal-2023-trust,10.5555/3618408.3619049,razeghi-etal-2022-impact}. %Even in environments resembling our real world, e.g. \tfont{\alfworld}, action sequences might be unintuitive and might not exactly correspond to our real world, possibly leading to knowledge conflicts between the models' parametric knowledge and task information specified in the environment description \mareike{EXAMPLE}.

\paragraph{Data availability}
Data availability dictates what form of learning can be applied to update agent parameters, or what extra steps are required in order to make a specific learning paradigm applicable, e.g. additional annotation of trajectories\footnote{A \emph{trajectory} of length $n$ is a sequence a$_1$, o$_1$, $\cdots$, a$_n$, o$_n$, i.e. actions $a$ and their corresponding observations $o$.}. The major learning paradigms for agents comprise ideas from online reinforcement learning (e.g. \citealp{webshop, scienceworld}), i.e. learning over time by interacting with the environment based on observing its behaviour \cite{sutton2018reinforcement}, supervised learning from gold trajectories (e.g. \citealp{apibank, toolbench, codeact, toolllm}), i.e. minimizing a loss function based on (dis)similarity between predicted trajectories and gold reference trajectories, and in-context learning (e.g. \citealp{webarena, visualwebarena, appworld}), i.e. prompting an in-context learner with a task description and possibly some example trajectories. 

While all tasks considered in this survey are generated based on interactive environments, %%SHORT, i.e. based on a well-defined transition function, 
not every dataset releases an interactive environment implementing the transition function. Some datasets (e.g. \tfont{\mindtoweb}, \tfont{\aitw}) only release task trajectories collected in the respective environment, i.e. sequences of actions and observations leading to task completion for a given goal. %, collected in the interactive environment, but not the environment itself. 
Without additional adjustments, such tasks are usually addressed via supervised or in-context learning. If datasets release interactive environments plus some explicit trajectories, either manually annotated (e.g. \tfont{\appworld}) or model-generated and verified for validity (e.g. \tfont{\toolbench}), such data can be used for supervised learning, or as few-shot examples for in-context learners.

\paragraph{Task generation}

%SHORT We consider as \textit{dataset} a collection of tasks collected for a similar environment (subsets of a common action, state, and observation space). 
Most task collections are created by crafting an action space and a transition function. Then, task instances are manually designed by coming up with a goal specification (e.g. invented by authors, crawled from the web, inspired by user surveys), and manual annotation of corresponding goal conditions. Some datasets provide \emph{problem generators}, which can automatically generate new solvable tasks, e.g. for training agents, or for evaluating agents on tasks with specific properties.  \tfont{\blocksworld} comes with a problem generator that given specific task properties, such as the number of blocks, automatically generates a corresponding pair of start state and goal conditions. \tfont{\appworld} provides template-specific task generators, i.e. given an instruction template like \exfont{I like
the last $\{$last-color$\}$ $\{$apparel$\}$ I bought on Amazon, repurchase the same in that size}, the generator instantiates the environment such that the task is solvable. %SHORT and automatically derives goal conditions. 

\section{Discussion and Future Directions}\label{sec:discussion}
In this survey, we structured the landscape of current tasks for developing and evaluating goal-directed interactive agents. The rapid progress towards such agents is exciting, and the breadth of tasks they are applied to indicates the huge impact agents will have on our daily lives once the technology is well-functioning and reliable.  We see several important directions for advancing agent technology that go beyond introducing more and more challenging environments.

\paragraph{Enabling agent-user interaction}

%This survey focuses on agents interacting with external environments, with no involvement of the user besides initially specifying their goal.\footnote{Environments simulating users for providing verbalized reward signals are discussed in \cref{sec:observations}.} 
The tasks focused on in this survey require no involvement of the user besides initially specifying their goal.\footnote{Environments simulating users for providing verbalized reward signals are discussed in \cref{sec:observations}.} 
%In more realistic scenarios users might, however, be  unable to provide full goal specifications up front,  %e.g. if final intents depend on information that only becomes available at later stages \mareike{CITE sth}, or 
In more realistic scenarios, providing full goal specifications up front might be infeasible, e.g. because users might be unable to explicitly verbalize their preferences \cite{lin2024decisionoriented}. Here, interaction between agent and user is required throughout the task completion process. 

\citet{lin2024decisionoriented} find that current agent architectures struggle with this for tasks requiring both user interaction and tool use, e.g. with asking relevant questions. We believe that combining goal-directed environment interaction with agent-user interaction, building on insights from scenarios for task-oriented dialogue (e.g. \citealp{multiwoz}) or collaborative games (e.g. \citealp{jeknic2024dialogue}), will render future agent technology more useful and realistic.

\paragraph{Targeted evaluation of agent behaviour}
%Targeted evaluation of specific agent skills is useful for \mareike{X}. 
Several works introduce dedicated data splits for evaluating specific aspects of agent behaviour, e.g. generalization to unseen actions \cite{toolbench, apibank} or compositional generalization \cite{compwob}. As end-to-end task completion in complex environments remains challenging, some works rather focus on separately addressing intermediate steps. For example, \citet{visualwebbench} focus on agents' understanding of websites, \citet{berkovitch2024identifyingusergoalsui} evaluate to what extent agents can identify user intents based on observing their GUI interactions. So far, failure cases are mainly reported through anecdotal error analysis. Formalizing these observations and developing frameworks for 
studying them in a targeted fashion will contribute to a 
better understanding of agent abilities and limitations.

\paragraph{Standardizing environments}

%The flood of new environments for evaluating agents holds the potential to advance agent research. To maximize usefulness, it is important that environments improve over time, i.e. new environments adjust for shortcomings observed in earlier benchmarks.

Future work on developing agent tasks should take into account shortcomings which have been observed for recent benchmarks. \citet{kapoor2024aiagentsmatter} identify several weaknesses in in existing benchmarks, including a lack of standardized held-out splits, and a lack of standardized evaluation scripts for reproducible evaluations. Another factor hindering reproducibility is dependence on external, i.e. not fully controllable, tools or APIs. For example, \tfont{\gaia} relies on a set of GPT4 plugins, which are constantly under development and subject to change. \tfont{\toolbench} relies on web-based APIs, and outputs of API calls might change over time. \citet{stabletoolbench} address this issue by introducing a stable version of \tfont{\toolbench} based on a virtual API server. We expect the question of how to render environments and evaluation reproducible to become even more relevant with current efforts to deploy agents in less restricted environments, where they can e.g. design their own action spaces \cite{cai2024large}.

\bibliography{tacl2021, custom}
\bibliographystyle{acl_natbib}

%\appendix
%\input{appendix}

\end{document}